\documentclass[10pt]{article}

\usepackage{amsmath}
\usepackage{amsthm}
\usepackage{amssymb}
\usepackage{authblk}
\usepackage{algorithm}
\usepackage{algpseudocode}
\usepackage{algorithmicx}
\usepackage{caption}
\usepackage{color}
\usepackage{comment}
\usepackage{float}
\usepackage[top=1in, bottom=1in, left=1in, right=1in]{geometry}
\usepackage{graphicx}
\usepackage[utf8]{inputenc}
\usepackage{natbib}
\usepackage[]{units}
\usepackage{times}

\newcommand{\jacob}{\mathbf{J}}

\newcommand{\cost}{\mathcal{L}}

\newcommand{\bmx}[0]{\begin{bmatrix}}
\newcommand{\emx}[0]{\end{bmatrix}}

\newcommand{\vect}[1]{\mathbf{#1}}
\newcommand{\vects}[1]{\boldsymbol{#1}}
\newcommand{\matr}[1]{\mathbf{#1}}

\newcommand{\cov}[0]{\operatorname{Cov}}

\newcommand{\vb}[0]{\vect{b}}

\newcommand{\vh}[0]{\vect{h}}

\newcommand{\vx}[0]{\vect{x}}

\newcommand{\vs}[0]{\vect{s}}

\newcommand{\vm}[0]{\vect{m}}

\newcommand{\mW}[0]{\matr{W}}

\newcommand{\mM}[0]{\matr{M}}

\newcommand{\mX}[0]{\matr{X}}

\newcommand{\mU}[0]{\matr{U}}

\newcommand{\mS}{\matr{S}}

\newcommand{\PP}[0]{\vects{\phi}}
\newcommand{\PS}[0]{\vects{\psi}}

\newcommand{\vmu}[0]{\vects{\mu}}
\newcommand{\vsigma}[0]{\vects{\sigma}}
\newcommand{\vepsilon}[0]{\vects{\epsilon}}

\makeatletter
\renewcommand\AB@affilsepx{, \protect\Affilfont}
\makeatother

\title{Spatio-temporal Dynamics of Intrinsic Networks in Functional Magnetic Imaging Data Using Recurrent Neural Networks}

\begin{document}
\author[1, 2]{R Devon Hjelm~\thanks{Authors made equal contributions}}
\author[3,4]{Eswar Damaraju~\textsuperscript{*}}
\author[5]{Kyunghyun Cho}
\author[6]{Helmut Laufs}
\author[3]{\\Sergey M Plis}
\author[3, 4]{Vince Calhoun}
\affil[1]{Montr\'eal Institute for Learning Algorithms}
\affil[2]{Microsoft Research}
\affil[3]{The Mind Research Network}
\affil[4]{The University of New Mexico}
\affil[5]{New York University}
\affil[6]{Schleswig University Hospital}

\maketitle
\abstract{
We introduce a novel recurrent neural network (RNN) approach to account for temporal dynamics and dependencies in brain networks observed via functional magnetic resonance imaging (fMRI).
Our approach directly parameterizes temporal dynamics through recurrent connections, which can be used to formulate blind source separation with a conditional (rather than marginal) independence assumption, which we call RNN-ICA.
This formulation enables us to visualize the temporal dynamics of both first order (activity) and second order\ (directed connectivity) information in brain networks that are widely studied in a static sense, but not well-characterized dynamically.
RNN-ICA predicts dynamics directly from the recurrent states of the RNN in both task and resting state fMRI.
Our results show both task-related and group-differentiating directed connectivity.
}

\section{Introduction}

Functional magnetic resonance imaging (fMRI) of blood oxygenation-level dependent (BOLD) signal provides a powerful tool for studying temporally coherent patterns in the brain \citep{damoiseaux2006, vince2008, smith2009}.
Intrinsic networks \citep[INs,][]{biswal1995} and functional connectivity are important outcomes of fMRI studies which illuminate our understanding of healthy and diseased brain function \citep{vince2001, allen2012b}.
While deep or nonlinear approaches for INs from fMRI and MRI exist \citep{hjelm2014restricted, plis2013deep, castro2016deep}, of the tools available, the most widely used are \emph{generative models} with shallow and linear structure.
Such models typically use a shared parameterization of structure to learn a common model across subjects which refactor the data into a constrained space that both provides straightforward analysis and allows for efficient and effective learning algorithms.

The most popular of such methods, independent component analysis \citep[ICA,][]{bell1995}, begins with the hypothesis that the data is a mixture of maximally independent sources.
ICA is trainable through one of many relatively simple optimization routines that maximize non-Gaussianity or minimize mutual information~\citep{hyvarinen2000independent}.
However, ICA, as with other popular linear methods for separating INs, is order-agnostic in time: each multivariate signal at each time step is treated as independent and identically distributed (i.i.d.).
While model degeneracy in time is convenient for learning; as an assumption about the data the explicit lack of temporal dependence necessarily marginalizes out dynamics, which then must be extrapolated in post-hoc analysis.

In addition, ICA, as it is commonly used in fMRI studies, uses the same parameterization across subjects, which allows for either temporal or spatial variability, but not both~\citep{calhoun2001spatial}.
The consequence of this is that ICA is not optimized to represent variation of shape in INs while also representing variation in time courses.
This may encourage ICA to exaggerate time course statistics, as any significant variability in shape or size will primarily be accounted for by the time courses.

Despite these drawbacks, the benefits of using ICA for separating independent sources in fMRI data is strongly evident in numerous studies, to the extent that has become the dominant approach for separating INs and analyzing connectivity \citep{zuo2010, kim2008hybrid, allen2012b, damoiseaux2006, vince2008, smith2009, calhoun2012multisubject}. 
In order to overcome shortcomings in temporal dynamics and subject/temporal variability, but without abandoning the fundamental strengths of ICA, we extend ICA to model sequences using recurrent neural networks (RNNs). 
The resulting model, which we call RNN-ICA, naturally represents temporal dynamics through a sequential ICA objective
and is easily trainable using back-propogation and gradient descent.

\section{Background}
Here we will formalize the problem of source separation with temporal dependencies and formulate the solution in terms of maximum likelihood estimation (MLE) and a recurrent model that parameterizes a conditionally independent distribution (i.e., recurrent neural networks).

Let us assume that the data is composed of $N$ ordered sequences of length $T$,
\begin{align}
\mX_n = (\vx_{1, n}, \vx_{2, n}, \dots, \vx_{T, n}),
\end{align}
where each element in the sequence, $\vx_{t, n}$, is a $D$ dimensional vector, and the index $n$ enumerates the whole sequence.
The goal is to find/infer a set of source signals,
\begin{align}
\mS_n = (\vs_{1, n}, \vs_{2, n}, \dots, \vs_{T', n}),
\end{align} 
such that a subsequence $\vs_{t_1:t_2} = (\vs_{t_1, n}, \vs_{t_1+1, n}, \dots, \vs_{t_2, n})$ \emph{generates} a subsequence of data, \\$\vx_{t'_1:t'_2} = (\vx_{t'_1, n}, \vx_{t'_1+1, n}, \dots, \vx_{t'_2, n})$, for $t_1 \leq t'_1 < t'_2$ and  $t_1 < t_2 \leq t'_2$.
In particular, we are interested in finding a generating function, 
\begin{align}
\mX_n = G(\mS_n, \vepsilon),
\label{eq:gen_eq}
\end{align}
where $\vepsilon$ is an additional noise variable.

This problem can generally be understood as \emph{inference} of unobserved or latent configurations from time-series observations.
It is convenient to assume that the sources, $\mS_n$, are stochastic random variables with well-understood and interpretable noise, such as Gaussian or logistic variables with independence constraints.
Representable as a directed graphical model in time, the choice of a-priori model structure, such as the relationship between latent variables and observations, can have consequences on model capacity and inference complexity.

Directed graphical models often require complex approximate inference which introduces variance into learning.
Rather than solving the general problem in Equation~\ref{eq:gen_eq}, we will assume that the generating function, $G(.)$, is \emph{noiseless}, and the source sequences, $\mS_n$ have the same dimensionality as the data, $\mX_n$, with each source signal being composed of a set of conditionally independent components with density parameterized by a recurrent neural network (RNN).
We will show that the learning objective closely resembles that of noiseless independent component analysis (ICA).
Assuming generation is noiseless and preserves dimensionality will reduce variance which would otherwise hinder learning with high-dimensional, low-sample size data, such as fMRI.

\subsection{Independent component analysis}
ICA~\citep{bell1995} hypothesizes that the observed data is a linear mixture of independent sources: $\vx_{t, n} = \sum_m s_{t, n, m} \vm_m$, where $\vs_{t, n} = \{s_{t, n, m}\}$ are sources and $\vm_m$ are the columns of a mixing matrix, $\mM$.
ICA constrains the sources (a.k.a., \emph{components}) to be maximally independent.
This framework presupposes any specific definition of component independence, and algorithms widely used for fMRI typically fall under two primary families, kurtosis-based methods and infomax~\citep{hyvarinen2000independent}, although there are other algorithms providing a more flexible density estimation~\citep{fu2014blind}.

For the infomax algorithm \citep{bell1995}, the model is parameterized by an unmixing matrix $\mW = \mM^{-1}$, such that $\mS_n = f(\mX_n) = \mW \cdot \mX_n$. 
In the context of fMRI, the infomax objective seeks to minimize the mutual information of $\vs_{n, t}$ for all subjects at all times.
This can be shown to be equivalent to assuming the prior density of the sources are non-Gaussian and that they factorize, or $p_s(\vs_{t, n}) = \prod_{m=1}^M p_{s_m}(s_{t, n, m})$, where $\vs_{t, n} = \{s_{t, n, m}\}$ is an $M$-dimensional vector.
When the sources are drawn from a logistic distribution, it can be shown that infomax is equivalent to maximum likelihood estimation (MLE), with the log-likelihood objective for the empirical density, $p_x(\mX_n)$, being transformed by $f(\mX) = \mW \cdot \mX$:
\begin{align}
\log p_x(\mX_n) = \log p_s(\mW \cdot \mX_n) + T \log |\det \mW|,
\end{align}
where $|\det \mW| = |\det \jacob_f (\mX)|$ is the absolute value of the determinant of the Jacobian matrix.

With ICA, generating example sequences can be done by applying the inverse of the unmixing matrix to an ordered set of sources.
However, one cannot simply sample from the model and generate samples of the observed data: any attempt to do so would simply generate unordered data and not true sequences.
The sources in ICA are constrained to be marginally independent in time; ICA does not explicitly model dynamics, and training on shuffled observed sequences will regularly produce the same source structure.

There are numerous graphical models and methods designed to model sequences, including hidden Markov models (HMMs) and sequential Monte Carlo \citep[SMC,][]{doucet2001introduction}.
HMMs are a popular and simple generative directed graphical models in time with tractable inference and learning and a traditional approach in modeling language.
However, HMMs place a high burden on the hidden states to encode enough long-range dynamics to model entire sequences.
Recurrent neural networks (RNNs), on the other hand, have the capacity to encode long-range dependencies through deterministic hidden units.
When used in conjunction to the ICA objective, the resulting algorithm is a novel and, as we will show, much more powerful, approach for blind source separation based on a conditional independence assumption.

\subsection{Recurrent neural networks}
\label{sec:RNN}
An RNN is a type of neural network with cyclic connections that has seen widespread success in neural machine translation \citep{cho2014learning}, sequence-to-sequence learning \citep{sutskever2014sequence}, sequence generation \citep{graves2013generating}, and numerous other settings.
When computing the internal state across a sequence index (such as time or word/character position), RNNs apply the same set of parameters (i.e., connective \emph{weights}) at each step.
This gives the model the properties of translational symmetry and directed dependence across time, which are desirable if we expect directed dependence with the same update rules across the sequence.
In addition, this makes RNN relatively memory-efficient, as one set of parameters are used across the sequence dimension.

RNNs have many forms, but we will focus on those that act as probabilistic models of sequences, i.e.:
\begin{align}
p(\mX) = p(\vx_{1:T}) = p(\vx_1) \prod_{t=2}^T p(\vx_t | \vx_{1:t-1}).
\end{align}
Better known as a ``language model" or \emph{generative} RNN, the exact form of the conditional density typically falls under a family of transformations,
\begin{align}
p(\vx_t | \vx_{1:t-1}) = f(\vh_t; \PS); \quad \vh_t = g(\vh_{t-1}, \vx_{t-1}; \PP),
\end{align}
where $\vh_t$ are a set of deterministic \emph{recurrent} states (or ``recurrent unit"). 
$g(.; \PP)$ are recurrent connections that take the current observation and hidden state as input and output the next recurrent state. 
The output connections, $f(.; \PS)$, take the recurrent states as input at each step and output the parameters for the conditional distribution.
Note that the model parameters, $\PS$ and $\PP$, are \emph{recurrent}: the same parameters are used at every time step and are not unique across the sequence index, $t$.

\begin{figure}[t]
\centering
\begin{minipage}{0.45\textwidth}
	\includegraphics[trim={0 9.5cm 8cm 0}, clip, scale=.2]{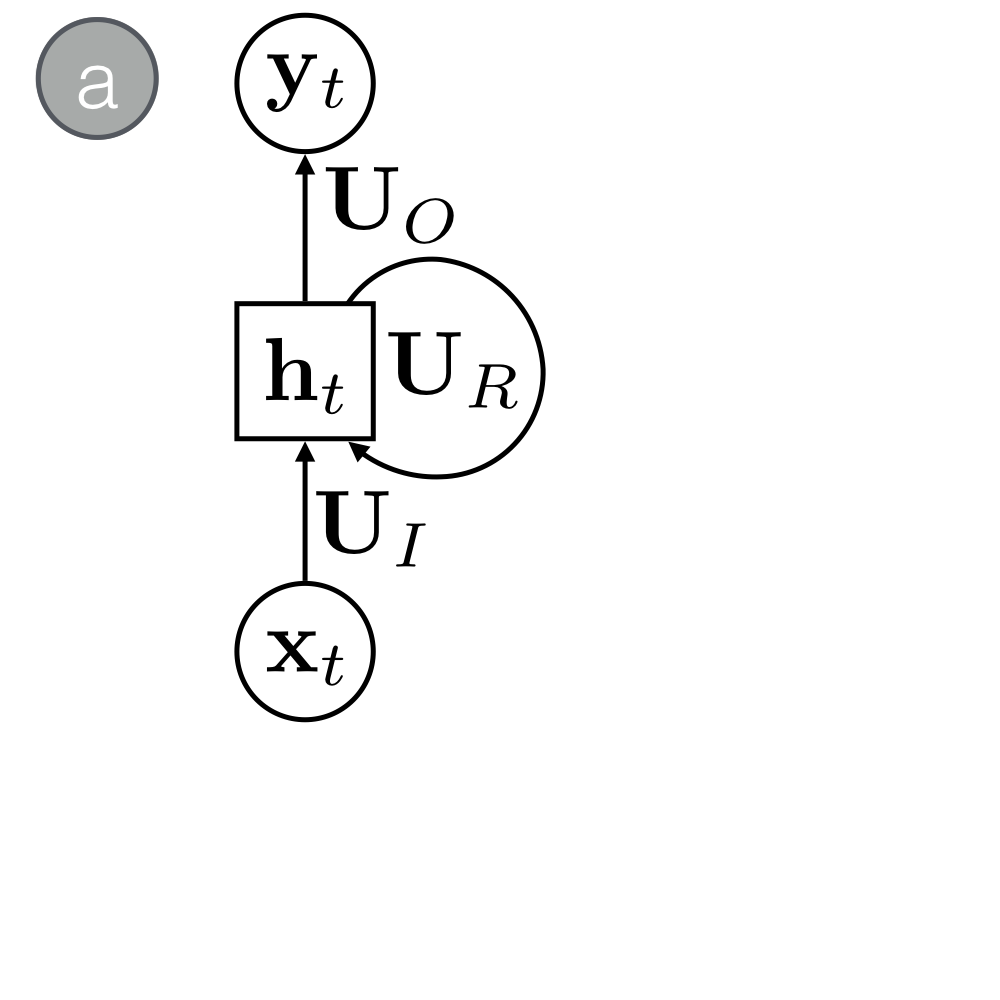}
\end{minipage}
\begin{minipage}{0.45\textwidth}
	\includegraphics[trim={0cm 4cm 8cm 0.5cm}, clip, scale=.2]{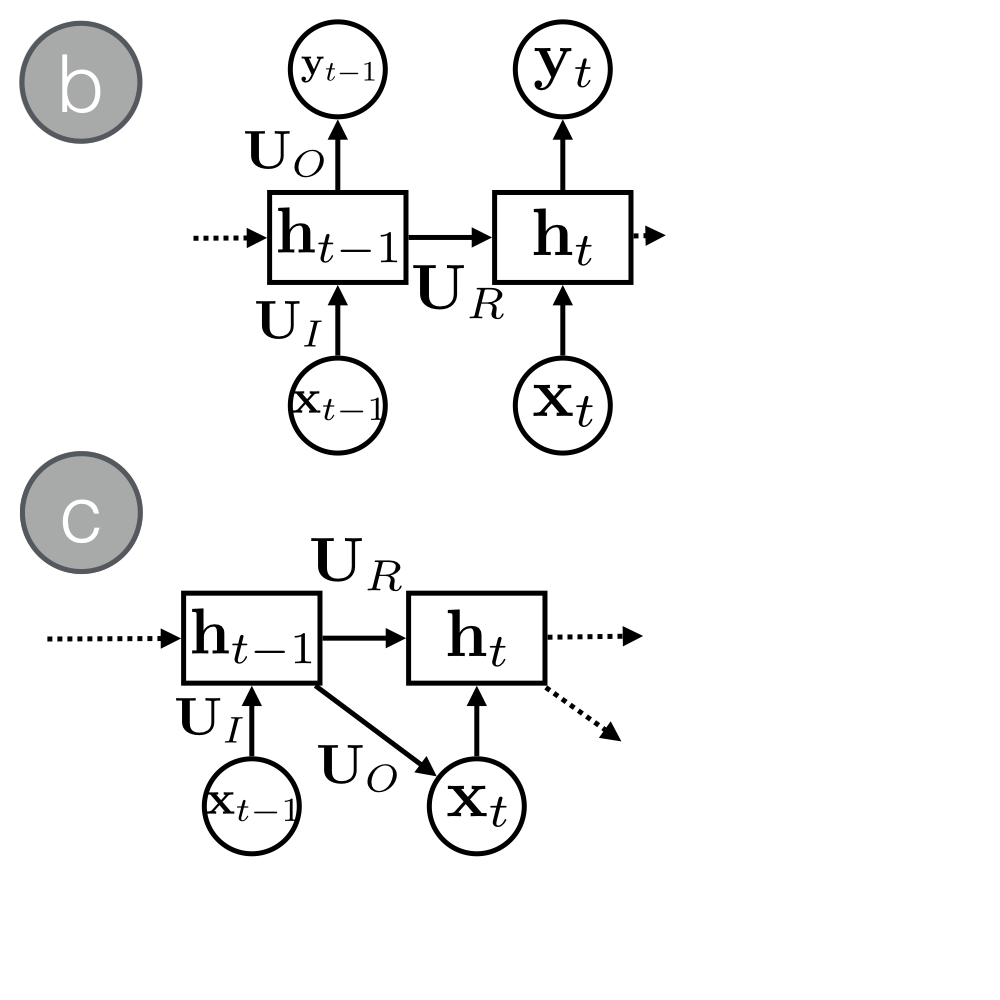}
\end{minipage}
\caption{a) a basic RNN with recurrent units $\vh_t$, recurrent connections, $\mU_R$, input connections, $\mU_I$, and output connections, $\mU_O$. b) An RNN with the recurrent connections rolled out. c) An RNN for sequence modeling and generation.}
\label{fig:rnn}
\end{figure}

The most canonical RNN for sequence modeling has the a simple parameterization (e.g., see Figure~\ref{fig:rnn}):
\begin{align}
g(\vh_t; \PS) = \tanh(\mU_R \vh_{t-1} + \mU_I \vx_{t-1} + \vb),
\label{eq:RNN}
\end{align}
where $\mU_R$ is a square matrix of recurrent weights, $\mU_I$ are the input weights, and $\vb$ is a bias term.
The mappings between the various variables in the model need not be shallow: an RNN with deep neural networks can model more complex recurrent transitions.
Parameterizations that use gating and other types of memory functions, such as long short-term memory \citep[LSTM,][]{hochreiter1997long} and gated recurrent units \citep[GRUs,][]{cho2014learning}, can be used to better model longer sequences and are also widely used.

Training an RNN for simple sequence modeling is easily done with the back-propagation algorithm, using the negative log-likelihood objective over the output conditional distributions:
\begin{align}
\cost = -\frac{1}{N} \sum_{n=1}^N \left( \sum_{t=2}^T \log p(\vx_t | \vx_{1:t-1}) + \log p(\vx_1) \right).
\end{align}

Typically the loss is computed with mini-batches instead of over the entire dataset for efficiency, randomizing mini-batches at each training epoch.
The marginal density, $p(\vx_1)$, can be learned by fitting to the average marginal across time, either to parameters of a target distribution directly or by training a neural network to predict the hidden state that generates $\vx_1$ \citep{bahdanau2014neural}.

\section{Methods}
\subsection{RNN-ICA}
RNNs have already been shown to be capable of predicting signal from BOLD\ fMRI data~\citep{gucclu2017modeling}, though usually in the supervised setting.
An unsupervised RNN framework for sequence modeling can easily be extended to incorporate the infomax objective. 
Define, as with ICA, a linear transformation for each observation to source configuration: $\vs_{t, n} = \mW \vx_{t, n}$, and define a high-kurtosis and factorized source distribution, $p_{s_{t, n}}(\vs_{t, n})$ (such as a logistic or Laplace distribution) for each time step, $t$, and each fMRI sequence, $n$.
We apply this transformation to an fMRI time series: $\vs_{1:T, n} = f(\vx_{1:T, n}) = (\mW \vx_{1, n}, \mW \vx_{2, n}, \dots \mW \vx_{T, n})$.
The log-likelihood function over the whole sequence, $\mX_n = \vx_{1:T, n}$, can be re-parameterized as:
\begin{align}
\log p(\vx_{1:T, n}) &= \log p(\vx_{1, n}) + \sum_{t=2}^{T} \log p(\vx_{t, n} | \vx_{1:t-1, n}) \nonumber\\
&= \log p_{s_1}(\mW \vx_{1, n}) + \sum_{t=2}^{T} \log p_{s_{t, n}}(\mW \vx_{t, n} | \vx_{1:t-1, n}) + \log |\det \jacob_f(\vx_{1:T, n})| \nonumber\\
&= T \log |\det \mW| + \log p_{s_{1, n}}(\mW \vx_{1, n}) + \sum_{t=2}^T \log p_{s_{t, n}}(\mW \vx_{t, n} | \vx_{1:t-1, n}),
\end{align}
where $\jacob_f$ is the Jacobian over the transformation, $f$, and the source distribution, $p_{s_{t, n}}$, has parameters determined by the recurrent states, $\vh_{t, n}$.
A high-kurtosis distribution is desirable to ensure independence of the sources (or minimizing the mutual information, e.g., the infomax objective~\citep{bell1995}), so a reasonable choice for the outputs of the RNN at each time step are the mean, $\vmu$, and scale, $\vsigma$, for a logistic distribution:
\begin{align}
\vmu_{t, n} = \mW_{\mu} \vh_{t, n};
\qquad
\vsigma_{t, n} = \mW_{\sigma} \vh_{t, n}.
\label{eq:pred}
\end{align}

\begin{figure}[t]
\centering
\includegraphics[scale=.45]{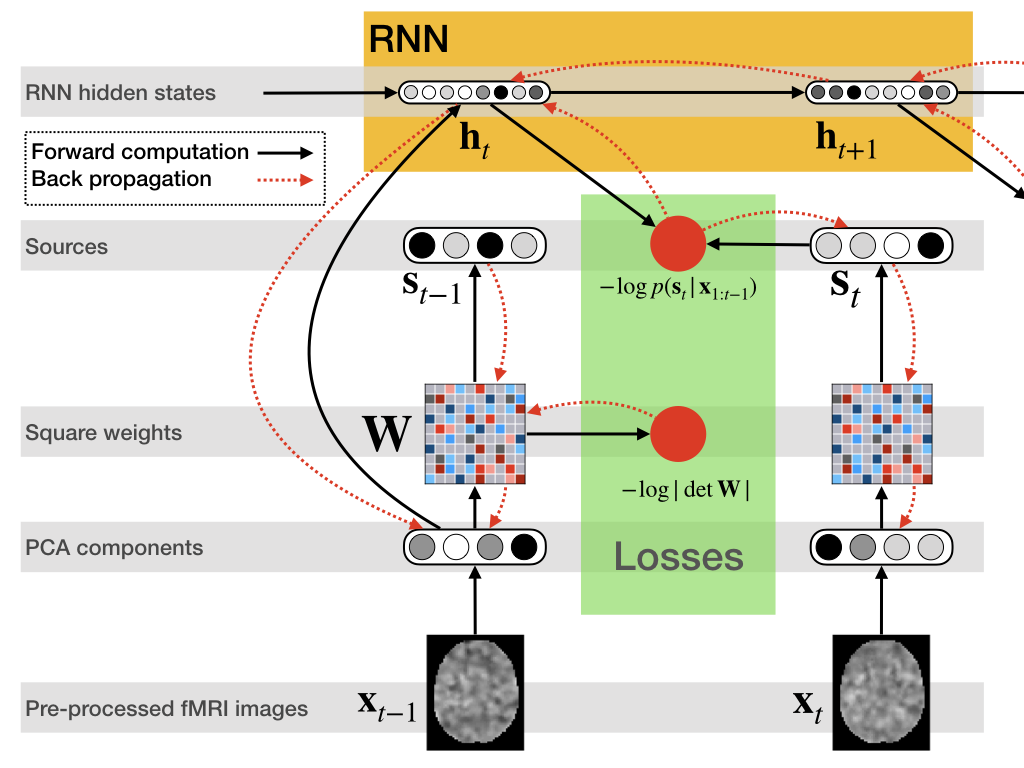}
\caption{
\small
RNN ICA. Preprocessed fMRI images are transformed and dimensionality-reduced using pre-trained PCA. The PCA components are passed through a square matrix which is the same for every subject and time-point. The PCA components are also passed as input to an RNN to compute the hidden states with the help of the previous state. These states are used to compute the likelihood of the next source in time. After the source time series is computed as well as likelihoods, the loss is back-propagated through the network for training.
}
\label{fig:rnn-ica}
\end{figure}

Figure~\ref{fig:rnn-ica} illustrates the network structure for a few time steps as well as the forward and back-propagated signal, and Algorithm~\ref{alg:rnnica} demonstrates the training procedure for RNN-ICA.
For our model, all network parameters and the ICA weight / un-mixing matrix, $\mW$, are the same for all subjects at all times.
Our treatment assumes the ICA weight matrix is square, which is necessary to ensure a tractable determinant Jacobian and inverse.
fMRI data is very high dimensional, so to reduce the dimensionality, we must resort to some sort of dimensionality reduction as preprocessing.
A widely used for dimensionality reduction in ICA studies of fMRI is principle component analysis (PCA) \citep{vince2001, allen2012b}, used to reduce the data to match the selected number of sources, $\vs_{t, n}$.

Note that RNNs with deeper architectures have been very successful for generative tasks~\citep[e.g., WaveNets, ][]{van2016wavenet}, and RNN-ICA could benefit from a deeper architecture capable of inferring more complex relationships in the data.
However, as fMRI data is often composed of a low number of training samples, we found it necessary to demonstrate the ability of RNN-ICA to learn meaningful sources with a simple RNN architecture.
We leave architectural improvements for RNN-ICA for future research.

\begin{algorithm}[t]
    \begin{algorithmic}
        \State $\mW \gets \text{initialize the unmixing matrix}$
        \State $\theta = (\mU_R, \mU_I, b) \gets \text{initialize the RNN recurrent and input weights and bias (see Equation \ref{eq:RNN})}$
        \State $\psi = (\mW_{\mu}, \mW_{\sigma}) \gets \text{initialize the RNN output weights (see Equation \ref{eq:pred})}$
        \State $\mathcal{D} = \{\vx_{1:T, n}\}_{n=1}^N \gets \text{N sequences of PCA-preprocessed fMRI sequences, windowed at $T$ time steps.}$
        \Repeat
        \State $\vx_{1:T, n} \sim \mathcal{D}$
        \Comment{Draw random samples from the set of PCA sequences}
        \State $\vs_{1:T,n} \gets (\mW \vx_{1, n}, \mW \vx_{2, n}, \dots \mW \vx_{T, n})$
        \Comment{Transform the PCA component sequence with the unmixing matrix}
        \State $\vh_{1,n} \gets f(\vx_{1, n})$ 
        \Comment{Initialize the first hidden state, as discussed in Section \ref{sec:RNN}}
        \State $(\vmu_{1, n}, \vsigma_{1, n}) \gets (\mW_{\mu} \vh_{1, n}, \mW_{\sigma} \vh_{1, n})$
            \Comment{Compute the parameters of the initial probability distribution}
        \For{\text{$t$ from $2$ to $T$}}
            \State $\vh_{t,n} \gets \tanh(\mU_R \vh_{t-1, n} + \mU_I \vx_{t-1, n} + \vb)$
            \Comment{Update each hidden state and conditional in-sequence}
            \State $(\vmu_{t, n}, \vsigma_{t, n}) \gets (\mW_{\mu} \vh_{t, n}, \mW_{\sigma} \vh_{t, n})$
            \Comment{Compute the parameters of the conditional probability at time $t$}
        \EndFor
        \State $\mathcal{L} \gets -T \log |\det \mW| - \log p_{s_{1, n}}( \vs_{1, n}) + \sum_{t=2}^T \log p_{s_{t, n}}(\vs_{t, n} | \vx_{1:t-1, n})$
        \Comment{Compute the negative log likelihood}
        \State $\phi = (\mW, \theta, \psi) \gets \phi - \gamma \nabla_{\phi} \mathcal{L}$
        \Comment{Perform gradient descent on the parameters}
        \Until{convergence}
    \end{algorithmic}
\caption{\label{alg:rnnica}. RNN-ICA}
\end{algorithm}

\section{Experiments and Results}

We first apply RNN-ICA to synthetic data simulated to evaluate the model performance and subsequently on real  functional magnetic imaging (fMRI) data.
FMRI analyses typically falls under two categories: task-based and resting state analysis.
Task experiments typically involve subjects being exposed to a time-series of stimulus, from which task-specific components can be extrapolated.
In the case of RNN-ICA, this should reveal task-related directed connectivity and spatial variability, in addition to the usual task-relatedness of activity from ICA.
Resting-state data is often used to confirm the presence of distinct and functional states of the brain. We chose a dataset resting state experiment that also had simultaneous Electroencephalography (EEG) from which ground-state subject neurobiological states could be derived. For RNN-ICA, we should be able to find a correspondence between predicted activation as defined in our model and changes in state. As a result, this should provide a means to prevent false positives or negatives when interpreting resting state network or inter-group differences owing to (systematically) different sleep stages present in their examined cohorts.
\subsection{Experiments with simulated data}
To test the model, we generated synthetic data using SimTB toolbox~\citep{erhardt2012simtb} in a framework developed to assess dynamics functional connectivity~\citep{allen2012b} and described in Figure 1 of \citet{LEHMANN2017635}. 
A total of $1000$ subjects corresponding to two groups of subjects, simulated healthy (SimHC) and simulated schizophrenia patients (SimSZ) were generated. 
A set of $47$ time courses were generated for each SimHC and SimSZ subject with the constraint that they have five states (covariance patterns) and a transition probability matrix per group that dictates state transitions derived from data from prior work on real data~\citep{damaraju2014dynamic}. 
The initial state probabilities were also derived from that work. A sequence of 480 time points with a TR of $2$ seconds were generated. 
A total of $1000$ subjects ($500$ per group) were generated of which first $400$ from each group were used during training and remaining $200$ samples were used in testing the model. 
The parameters of hemodynamic response model (delay, undershoot etc) used to simulate the data were also varied per subject to introduce some heterogeneity. The known initial state of a subject and a transition probability matrix that governs transitions ensured a ground truth state transition vector (a vector of transitions between five simulated states unique to each subject).

\begin{figure}[t]
\centering
\includegraphics[scale=2]{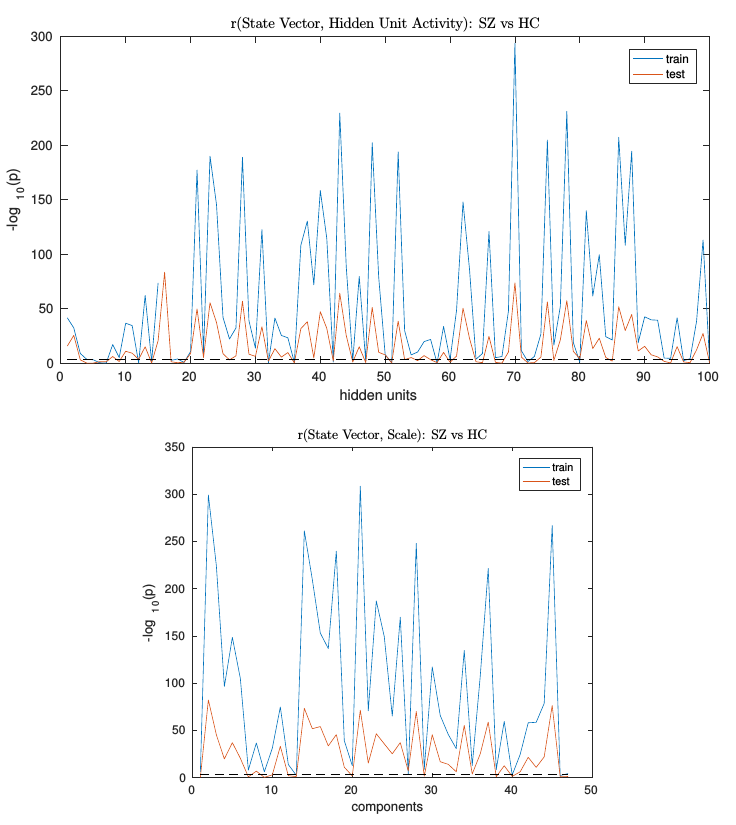}
\caption{
\small
The SimHC versus SimSZ group differences of correlation between RNN hidden unit activations (A) and component scale factors (B) to the ground truth state vectors for train (blue) and test (red) subjects. Shown are $-\log_{10}(p)$ values. Note same hidden units or component scale factors track group differences in both train and test cases although with lower strength in test cases. The dashed black line corresponds to the false discovery rate threshold of $0.001$.
}
\label{fig:sim_GrpDiff}
\end{figure}

An RNN-ICA model was then trained on the $800$ subject training data for $500$ epochs with model parameters similar to those in subsequent sections. The resultant sources $S$, the source distributions predicted by RNN ($\vmu$, and $\vsigma$), and the RNN hidden unit activations for each subject were then correlated to the subject’s ground truth state vector. 
The trained model was then run on the test data and correlations were again computed between their model outputs and state vectors. 
We then computed group differences between the correlation distributions of SimHC and SimSZ groups and are summarized for both training and test cases in Figure~\ref{fig:sim_GrpDiff}).
Our results show that RNN-ICA generalized group differences well to the test set in this setting, as represented in the hidden state activations and scaling factor.

\subsection{Task experiments}
To demonstrate the properties and strengths of our model, we apply our method to task fMRI data.
Data used in this work is comprised of task-related scans from $28$ healthy participants and $24$ subjects diagnosed with schizophrenia, all of whom gave written, informed, Hartford hospital and Yale IRB approved consent at the Institute of Living and were compensated for their participation.
All participants were scanned during an auditory oddball task (AOD) involving the detection of an infrequent target sound within a series of standard and novel sounds.
More detailed information regarding participant demographics and task details are provided in \cite{swanson2010}.

Scans were acquired at the Olin Neuropsychiatry Research Center at the Institute of Living/Hartford Hospital on a Siemens Allegra 3T dedicated head scanner equipped with \unitfrac[40]{mT}{m} gradients and a standard quadrature head coil.
The functional scans were acquired trans-axially using gradient-echo echo-planar-imaging with the following parameters: repeat time (TR) \unit[1.50]{s}, echo time (TE) \unit[27]{ms}, field of view 24 cm, acquisition matrix $\unit[64 \times 64]{}$, flip angle $70^\circ$, voxel size $\unit[3.75 \times 3.75 \times 4]{mm^3}$, slice thickness \unit[4]{mm}, gap \unit[1]{mm}, 29 slices, ascending acquisition.
Six ``dummy" scans were acquired at the beginning to allow for longitudinal equilibrium, after which the paradigm was automatically triggered to start by the scanner.
The final AOD dataset consisted of $249$ volumes for each subject.

Data underwent standard pre-processing steps using the SPM software package \cite[see][for further details]{vince2008}.
Subject scans were masked below a global mean image then each voxel was variance normalized.
Each voxel timecourses was then detrended using a 4th-degree polynomial fit, and this was repeated for all subjects.
PCA was applied to the complete dataset without whitening, and the first $60$ components were kept to reduce the data.
Finally, each PCA component had its mean removed before being entered into the model.

\subsubsection{Model and setup}
For use in RNNs, the data was then segmented into windowed data, shuffled, and then arranged into random batches.
Each PCA loading matrix for subject was comprised of $60$ PCA time courses of length $249$.
These were segmented into $228$ equal-length windowed slices using a window size of $20$ and stride of $1$.
The number of components roughly corresponds to the number found in other studies~\citep{vince2001, allen2012b, vince2008}, and $20$ time steps is equivalent to $30$ seconds, which has been shown provides a good trade-off in terms of capturing dynamics and not being overly sensitive to noise~\citep{vergara2017effect}.
The final dataset was comprised of $228$ volumes for each of the $52$ subjects with $60$ pca time courses each.
These were then randomly shuffled at each epoch into batches of $100$ volumes each from random subjects and time points.

We used a simple RNN with $100$ recurrent hidden units and a recurrent parameterization as in Equation \ref{eq:RNN}, as we do not anticipate needing to model long range dependencies that necessitate gated models \citep{hochreiter1997long}.
The initial hidden state of the RNN was a $2$-layer feed forward network with $100$ softplus $\left(\log(1 + \exp(x))\right)$ units using $20\%$ dropout.
An additional $L_2$ decay cost,  $\lambda \sum_{i, j}W_{i, j}^2$, was imposed on the unmixing matrix, $\mW$, for additional regularization with a decay rate of $\lambda=0.002$.
The model was trained using the RMSProp algorithm \citep{Hinton-Coursera2012} with a learning rate of $0.0001$ for $500$ epochs.

\subsubsection{Results}
\begin{figure}[H]
\centering
\includegraphics[trim={0cm 15cm 0cm 0}, scale=.35]{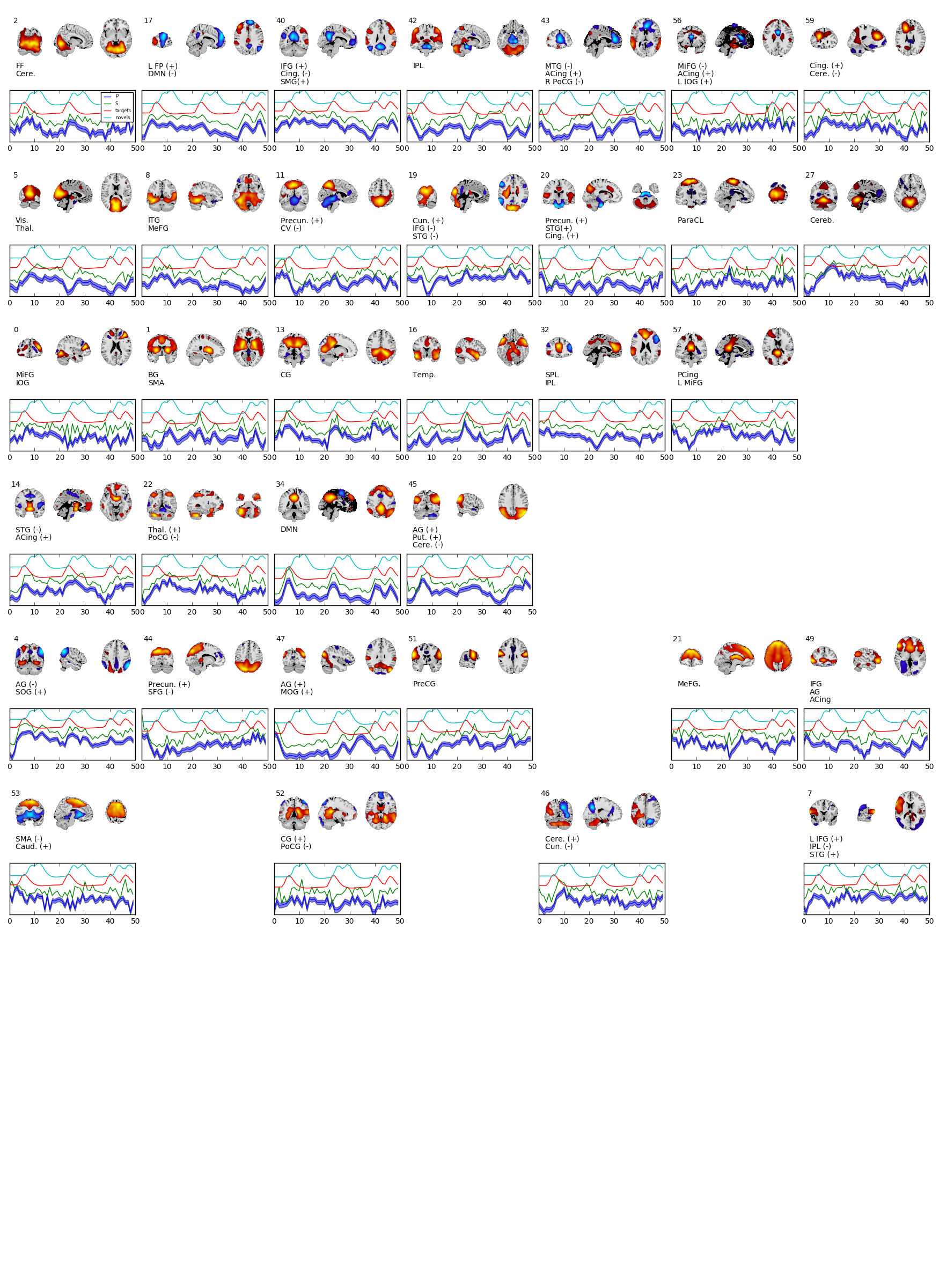}
\caption{
\small
Selected set of spatial maps from RNN-ICA without spatial map corrections.
Maps were filtered by hand, omitting grey matter, ventricle, and motion artifact features.
Source (green), mean-predicted with uncertainty (blue), and target (red) and novel (cyan) stimulus time courses are shown, each normalized to their respective variance and offset for easier visualization.
Each map was thresholded at 2 standard deviations and grouped according to FNC (see Figure~\ref{fig:UC_fnc}).
The spatial maps were sign flipped along with their respective time courses to ensure the distribution of back-reconstructed voxels had a positive skew.
The truncated ROI labels were found by visual inspection with the aid of the AFNI package~\citep{cox1996afni} and correspond to: 
MiFG: middle frontal gyrus, 
MeFG: medial frontal gyrus, 
SMeFG: superior medial frontal gyrus, 
IFG: inferior frontal gyrus, 
MOrbG: middle orbital gyrus, 
IPL: inferior parietal lobule, 
SPL: superior parietal lobule, 
IOG: inferior occipital gyrus, 
MOG: middle occipital gyrus, 
SOG: superior occipital gyrus, 
ITG: inferior temporal gyrus, 
STG: superior temporal gyrus, 
SMG: supramarginal Gy, 
PoCG: postcentral gyrus, 
PreCG: precentral gyrus, 
ParaCL: paracentral Lob, 
MCing: middle cingulate, 
ACing: anterior cingulate, 
PCing: posterior cingulate, 
AG: angular gyrus, 
BG: basal ganglia, 
SMA: supplementary motor area, 
FF: fusiform gyrus, 
CV: cerebellar vermis, 
CG: calcarine gyrus, 
FP: frontoparietal, 
DMN: default-mode network, 
ParaG: parahippocampal gyrus, 
LingG: lingual gyrus, 
WM: white matter, 
GM: white matter, 
Precun.: precuneus, 
Thal.: thalamus, 
Vis.: visual, Temp.: temporal, 
Cere.: cerebellum, 
Cun.: cuneus, 
Puta.: putamen,
Cing.: cingulate,
Caud.: caudate,
Pari.: parietal,
Front.: frontal,
Ins: insula,
Vent.: ventricle.
}
\label{fig:maps_UC}
\end{figure}

\begin{figure}[t]
\centering
\includegraphics[trim={2.5cm 4.5cm 4cm 2cm}, clip, scale=.45]{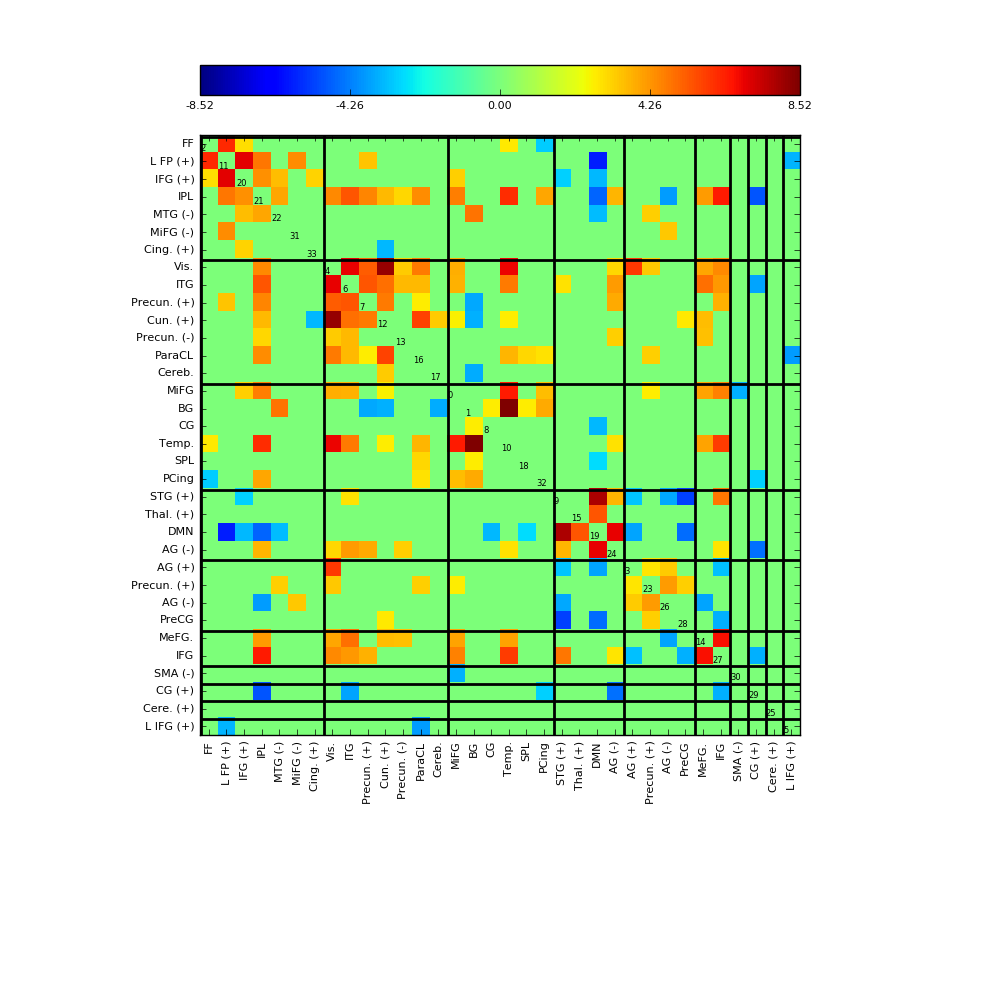}
\caption{
\small
Functional network connectivity~\citep[FNC,][]{jafri2008method} matrix, which is essentially the temporal cross correlation matrix, in this case averaged across subjects.
Grouping were found using a multi-level community algorithm~\cite{blondel2008fast}, ordering the FNC axes according the groups.
(+) and (-) in the labels indicate regions of interest which have majority positive and negative voxel values, respectively.
}
\label{fig:UC_fnc}
\end{figure}

\begin{table*}[th]
\scriptsize
\center
\caption{\small p-values from a 1-sample t-test over beta-values with $p\leq 10^{-7}$ for the target and novel stimulus for the sources, $\vs_{t, n}$, the predicted means, $\vmu_{t, n}$, and the predicted scale-factor, $\vsigma_{t, n}$.
Beta-values were found for each subject and component using multiple regression to target and novel stimulus, and t-tests were performed for each component over all subjects.
Among the most-significant task-related components to target stimulus include the middle temporal gyrus, default mode network, and the parietal lobule.
A legend for ROI label names can be found in the caption of Figure~\ref{fig:maps_UC}.
The (+/-) in the label name specify the sign of the map in Figure~\ref{fig:maps_UC}, while the (+/-) in the p-values specifies the sign of the corresponding t-value.
}
\begin{tabular}{| c | c | c | c | c | c | c |}
\hline
    &             & $\vmu_i$      &         & $\vs_i$       &         & $\vsigma_i$   \\
\hline
 ID & Label       & Targets & Novels  & Targets & Novels  & Targets \\
 \hline
 2  & FF          		& 3.1e-13 (+) 	&         		& 2.1e-08 (+) 	&         &         \\
 13 & CG          		& 2.2e-13 (+) 	&         		& 1.5e-11 (+) 	&         &         \\
 14 & STG (-)    	& 4.9e-08 (-) 	&         		& 1.4e-08 (-) 	&         &         \\
 16 & Temp.      		& 2.2e-10 (+) 	&         		& 1.9e-17 (+) 	&         &         \\
 17 & L FP (+)   	& 2.6e-11 (+) 	&         		&         &         	&         \\
 20 & Precun. (+) 	&         		&         		& 8.5e-09 (+) 	&         &         \\
 21 & MeFG.      	& 4.1e-08 (+) 	&         		&         &         	&         \\
 34 & DMN         	& 1.0e-18 (-) 	&         		& 1.3e-18 (-) 	&         & 1.8e-10 (+) \\
 42 & IPL         		& 4.7e-19 (+) 	& 4.4e-08 (+) 	& 1.3e-14 (+) 	& 1.1e-08 (+) &         \\
 43 & MTG (-)     	& 2.1e-19 (+) 	& 7.0e-15 (+) 	& 4.4e-15 (+) 	& 1.2e-14 (+) &         \\
 45 & AG (+)      	& 3.9e-09 (-) 	&         		&         		& 9.4e-08 (-) &         \\
 46 & Cere. (+)   	&         		& 5.4e-10 (+)	&         		&         &         \\
 47 & AG (+)      	& 1.4e-12 (-) 	& 4.5e-10 (+) 	& 6.6e-13 (-) 	& 2.0e-11 (-) &         \\
 57 & PCing       	&         		&        		&         		& 1.7e-09 (-) &         \\\hline
\end{tabular}
\label{tab:UC}
\end{table*}

Figure~\ref{fig:maps_UC} shows $34$ spatial maps back-reconstructed.
The spatial maps were filtered from the original $60$, omitting white matter, ventricle, and motion artifact features.
Each of the spatial maps along with their respective time-courses were sign-flipped to ensure that each back-reconstructed distribution of voxels had positive skew.
The maps are highly analogous to those typically found by linear ICA~\citep{vince2001, allen2012b}, though with more combined positive/negative features in one map.

Figure~\ref{fig:UC_fnc} shows the functional network connectivity \citep[FNC,][]{vince2001, jafri2008method} matrix, in which the components are grouped according to a multi-level community algorithm~\cite{blondel2008fast} using the symmetric temporal cross-correlation matrix.
For each subject and component, we performed multiple linear regression of the sources, $\vs_{t, n}$, the predicted means, $\vmu_{t, n}$, and the predicted scale-factor, $\vsigma_{t, n}$ from each subject to the target and novel stimulus.
Table~\ref{tab:UC} shows the p-values from a 1-sample t-test on the beta values across subjects for components with values of $p \leq 10^{-7}$.
Many components show similar task-relatedness across the source time courses and predicted means, notably temporal gyrus features, parietal lobule, and the default mode network (DMN, which is negatively correlated).
In addition, the DMN shows the strongest task-relatedness in the scale factor.

\begin{table*}[t]
\centering
\begin{minipage}[l]{0.7\textwidth}
\includegraphics[scale=.4]{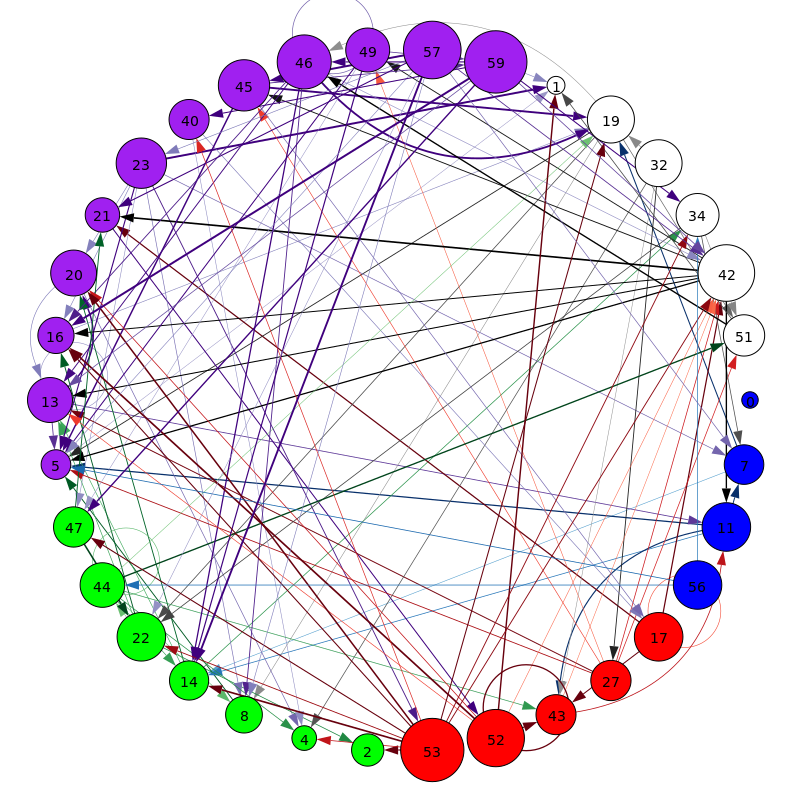}
\end{minipage}
\begin{minipage}[r]{0.2\textwidth}
\includegraphics[trim={1cm 1cm 5.7cm 1cm}, clip, scale=.7]{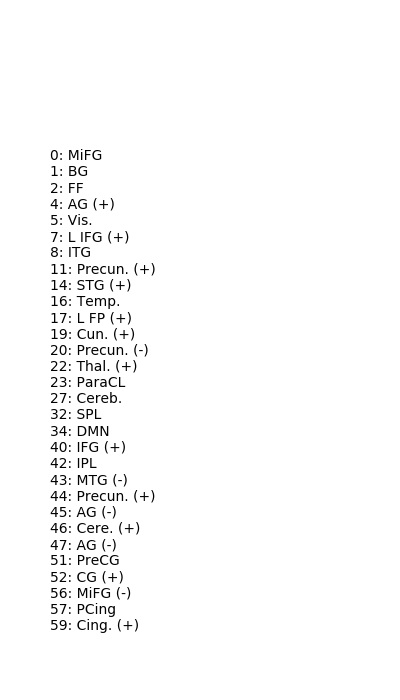}
\end{minipage}
\captionof{figure}{
\small
A graphical representation of the next-time Jacobian terms, $|\frac{\partial{\mu_{i, t}}}{\partial{s_{j, t-1}}}|$, averaged over time and subjects.
The features were grouped by a multi-level community algorithm~\cite{blondel2008fast}, using the Pearson correlation coefficient to define an undirected graph (see Equation \ref{eq:corr}). Corresponding ROIs are provided on the right, and the complete legend can be found in Figure~\ref{fig:maps_UC}.
Grouping (and coloring) was done by constructing an undirected graph using the Pearson coefficients, clustering the vertices using a standard community-based hierarchical algorithm.
}
\label{fig:graph_jacob}
\end{table*}

\begin{figure}
\centering
\includegraphics[trim={5cm 25cm 25cm 0}, scale=.4]{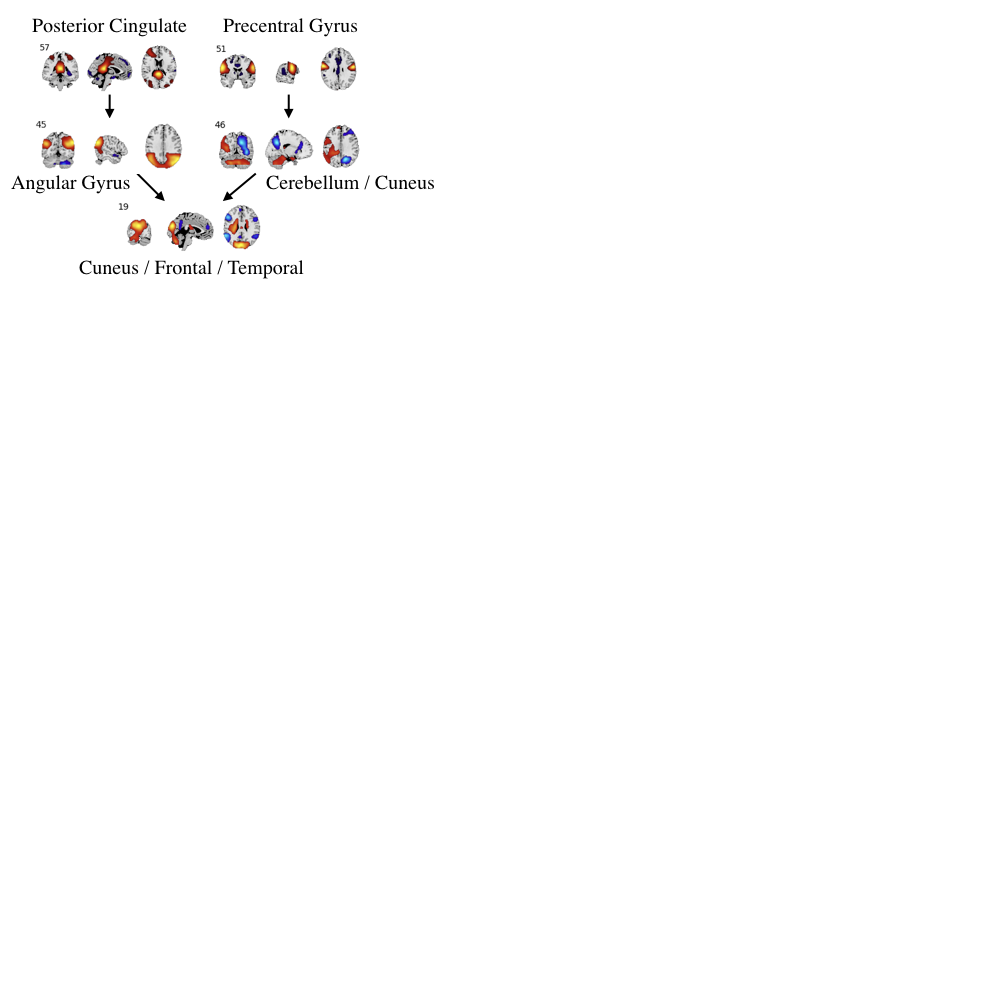}
\caption{
\small
An example of directed connectivity in task data derived from the Jacobian, as represented by the spatial maps.}
\label{fig:causal}
\end{figure}

In order to analyze how the RNN encodes dynamics, we analyze the Jacobian of the predicted mean of each component $i$ at time $t$ over all components at previous times, $t'$:
\begin{align}
\frac{\partial{\mu_{i, t}}}{\partial{s_{j, t'}}} = \sum_{k}\frac{\partial{x_{k, t'}}}{\partial{s_{j, t'}}}\frac{\partial{\mu_{i, t}}}{\partial{x_{k, t'}}}.
\end{align}
The derivatives are tractable, as the means, $\mu_{i, t}(\vx_{0:t-1})$, are differentiable functions w.r.t the input $\vx_{0:t-1}$.
These derivatives can can be interpreted as being a measure of \emph{directed connectivity} between components in time, as they represent the predicted change of a future component (as understand through the change of its mean value) given change of a previous component.
While the full Jacobian provides directed connectivity between source between all pairs of time, $(t, t')$, to simplify analysis, we only looked at next-time terms, or $t'=t-1$.

A representative graph is given in Figure~\ref{fig:graph_jacob}, where the thickness of the edges represents the strength of the directed connection as averaged across time and subjects with the sign removed ($|\frac{\partial{\mu_{i, t}}}{\partial{s_{j, t'}}}|$).
The color/grouping of the nodes corresponds to the similarity in directed connectivity as measured by the Pearson correlation coefficient:
\begin{align}
\rho_{i, j} =  \frac{\cov{\left(\bar{\nu}_i , \bar{\nu}_j\right)}}{\sigma_{\bar{\nu}_i} \sigma_{\bar{\nu}_j}}, \
\text{where} \ \bar{\nu}_i = \frac{1}{T}\sum_t \frac{\partial{\mu_{k, t}}}{\partial{s_{i, t}}},
\label{eq:corr}
\end{align}
$\cov{(.,.)}$ is the covariance, and $\sigma_{\bar{\nu}_i}$ is the standard deviation across the components indexed by $k$.
Grouping was done by constructing an undirected graph using the Pearson coefficients, clustering the vertices using the same community-based hierarchical algorithm as with the FNC above.
An example directed connectivity graph with the spatial maps is given in Figure~\ref{fig:causal}.

\begin{figure}[t]
\includegraphics[scale=0.4]{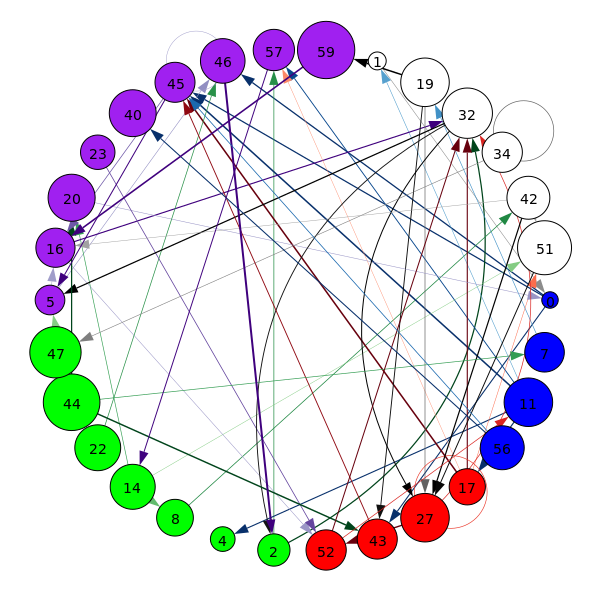}
\hspace{1cm}
\includegraphics[scale=0.4]{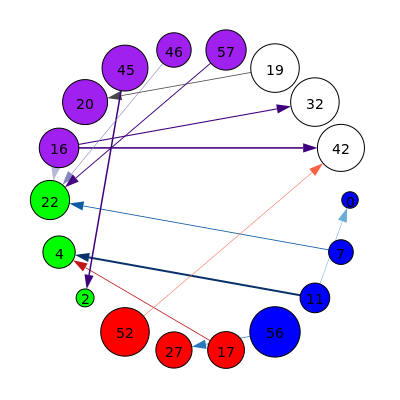}
   \caption{
      \small
A graphical representation of target (left) and novel (right) task-significant next-time Jacobian terms (see Figure~\ref{fig:graph_jacob} on grouping).
Target stimulus directed connectivity were thresholded at $p \leq 10^{-10}$, while novel directed connectivity where thresholded at $p \leq 10^{-7}$.
Target and novel graphs were thresholded at different values for cleaner graphical representations.
Legend for nodes is in Figure~\ref{fig:graph_jacob}.
      }
\label{fig:caus}
\end{figure}

\begin{figure}[t]
\centering
\includegraphics[trim={0cm 20cm 25cm 0cm}, clip, scale=0.4]{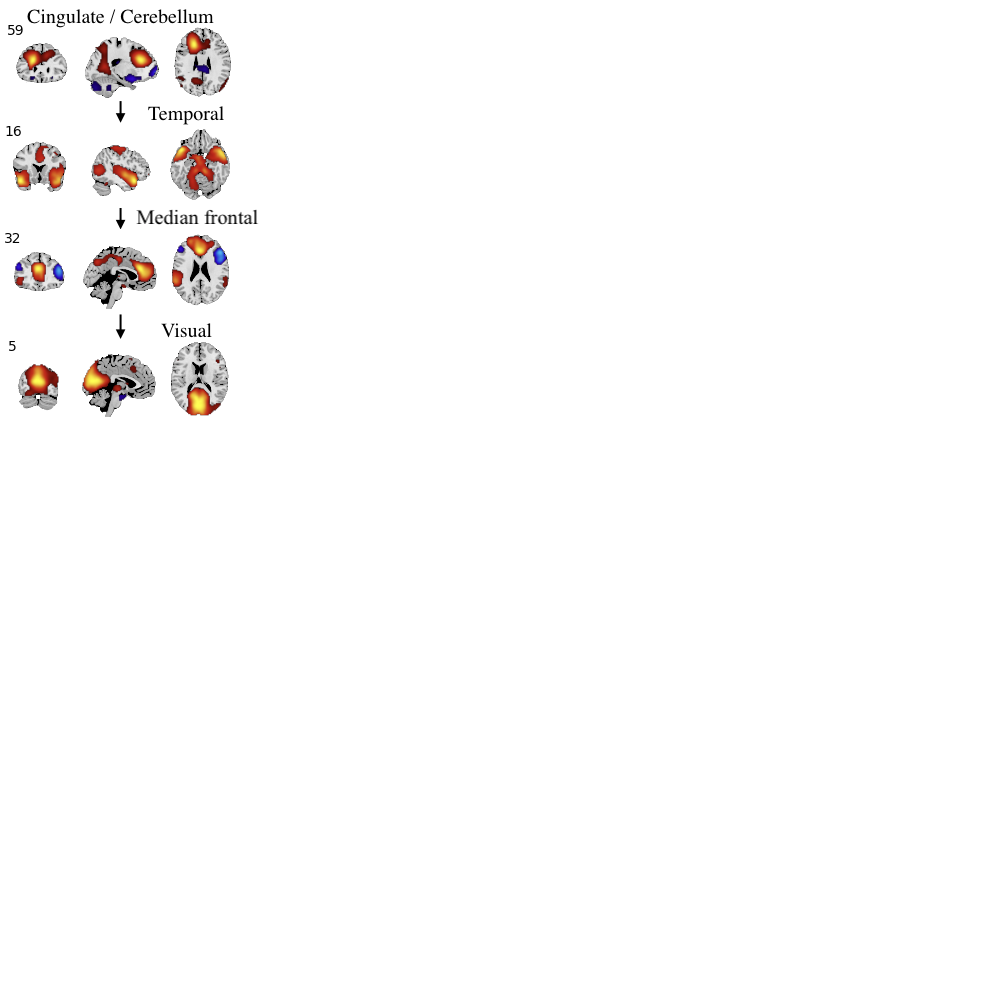}
\hspace{1cm}
\includegraphics[trim={0cm 20cm 15cm 0cm}, clip, scale=0.4]{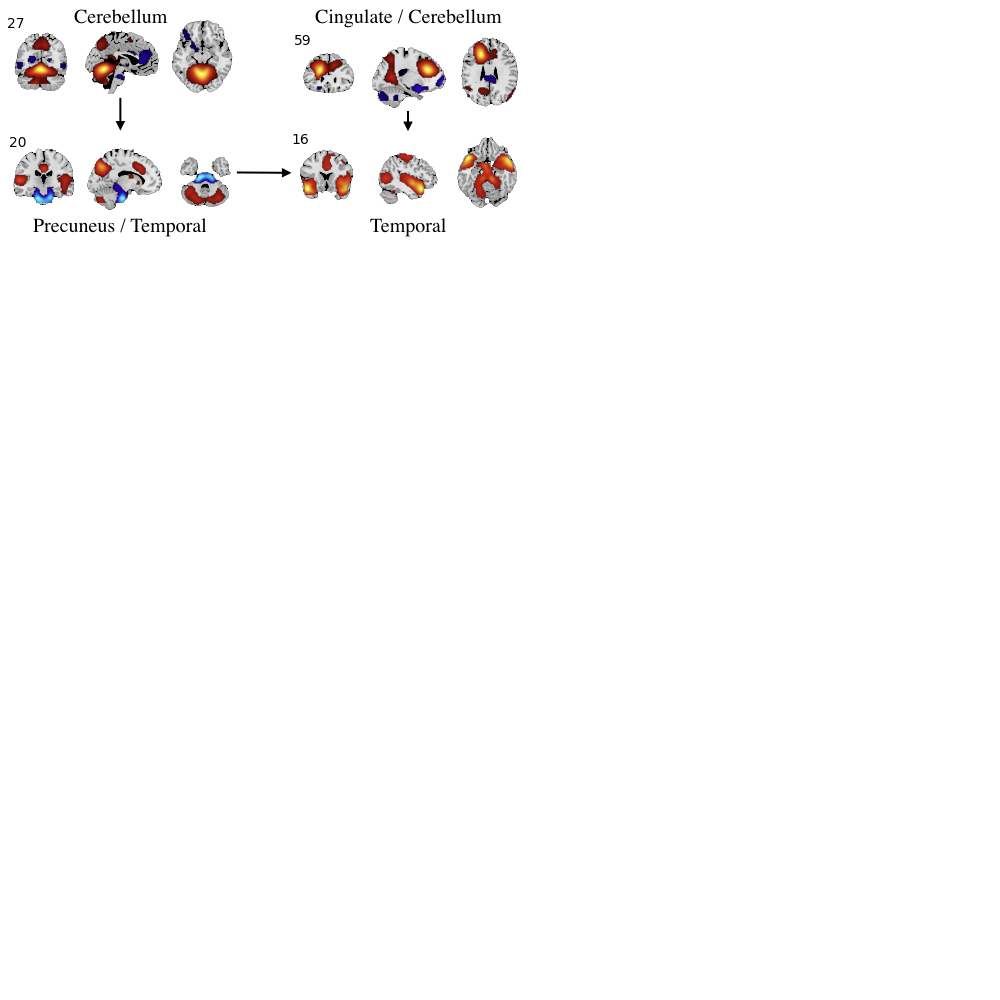}
\caption{
\small
An example of left: task-related (target stimulus) and right: group-differentiating causal relationships derived from the Jacobian, as represented by the spatial maps.}
\label{fig:causal_task}
\end{figure}

\begin{figure}[t]
\centering
  \includegraphics[scale=.45]{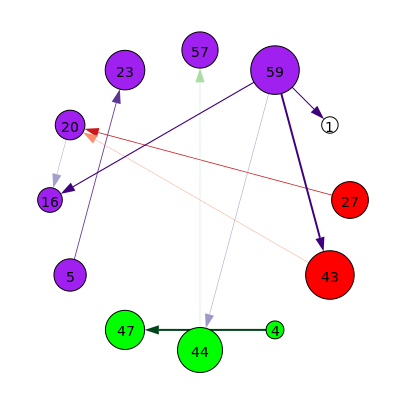}
\hspace{1cm}
  \includegraphics[scale=.45]{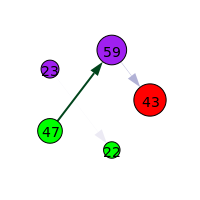}
   \caption{
      \small
A graphical representation of target (left) and novel (right) group-differentiating next-time Jacobian terms (see Figure~\ref{fig:graph_jacob} on grouping).
Target stimulus directed connectivity were thresholded at $p \leq 0.001$.
Legend for nodes is in Figure~\ref{fig:graph_jacob}.
This shows that the influence between components across time when different stimulus is present can vary across groups.
      }
   \label{fig:caus_d}
\end{figure}

Each of the next-step Jacobian terms were used as time-courses with a multiple-regression to target and novel stimulus, with significance tested using a one-sample t-test as with the time courses and a two-sample t-test across groups.
The resulting task-related directed connectivity are represented in Figure~\ref{fig:caus} for both targets and novels, with an example graph with spatial maps presented in Figure~\ref{fig:causal_task}.
Group-differentiating relationships are given in Figure~\ref{fig:caus_d} with an example graph with spatial maps given in Figure~\ref{fig:causal_task}.

\subsection{Resting state experiments}

We evaluated our model on resting state data to show RNN-ICA as a viable model and to demonstrate that properties of the network correspond to wake/sleep states.
Resting state functional MRI data was collected from $55$ subjects for $50$ minutes each ($1505$ volumes, TR= \unit[2.08]{s}) with a Siemens 3T Trio scanner while the subjects transitioned from wakefulness to at most sleep stage N3~\citep[see,][for more details]{TAGLIAZUCCHI201263}. 
This data was approved by ethics committee of Goethe University.
Simultaneous EEG was acquired facilitating sleep staging per AASM criteria resulting in a hypnogram per subject (a vector assignment of consecutive \unit[30]{s} EEG epochs to one of wakeful(W), N1, N2 and N3 sleep stae). 
We discarded first $5$ time points to account for T1 equilibration effects.

After performing rigid body realignment and slice-timing correction, subject data was warped to MNI space using SPM12. 
Then voxel time courses were despiked using AFNI. 
We then regressed out voxel time courses with respect to their head motion parameters (and their derivatives and squares), their mean white matter and CSF signals. 
Next, we bandpass filtered the data with a passband of \unit[0.01 - 0.15]{Hz}. 
We extracted mean ROI time courses from $268$ nodes extracted from the bioimage suite~\citep{papademetris2006bioimage} and reported in \cite{shen2013groupwise}. 

\subsubsection{Model and setup}
We used the same model and training procedure as with our task data analysis in the previous section. Of the 55 subjects, 50 subjects were used during training and 5 subjects were left out for testing.
We then examined the correspondence between hidden recurrent units of the trained model and subject hypnogram as well as between mean and scale of predictive source distribution and hypnogram. Similar tests were run on the model outputs on the 5 left out test cases.

\subsubsection{Results}

\begin{figure}[t]
\centering
\includegraphics[width=0.9\textwidth]{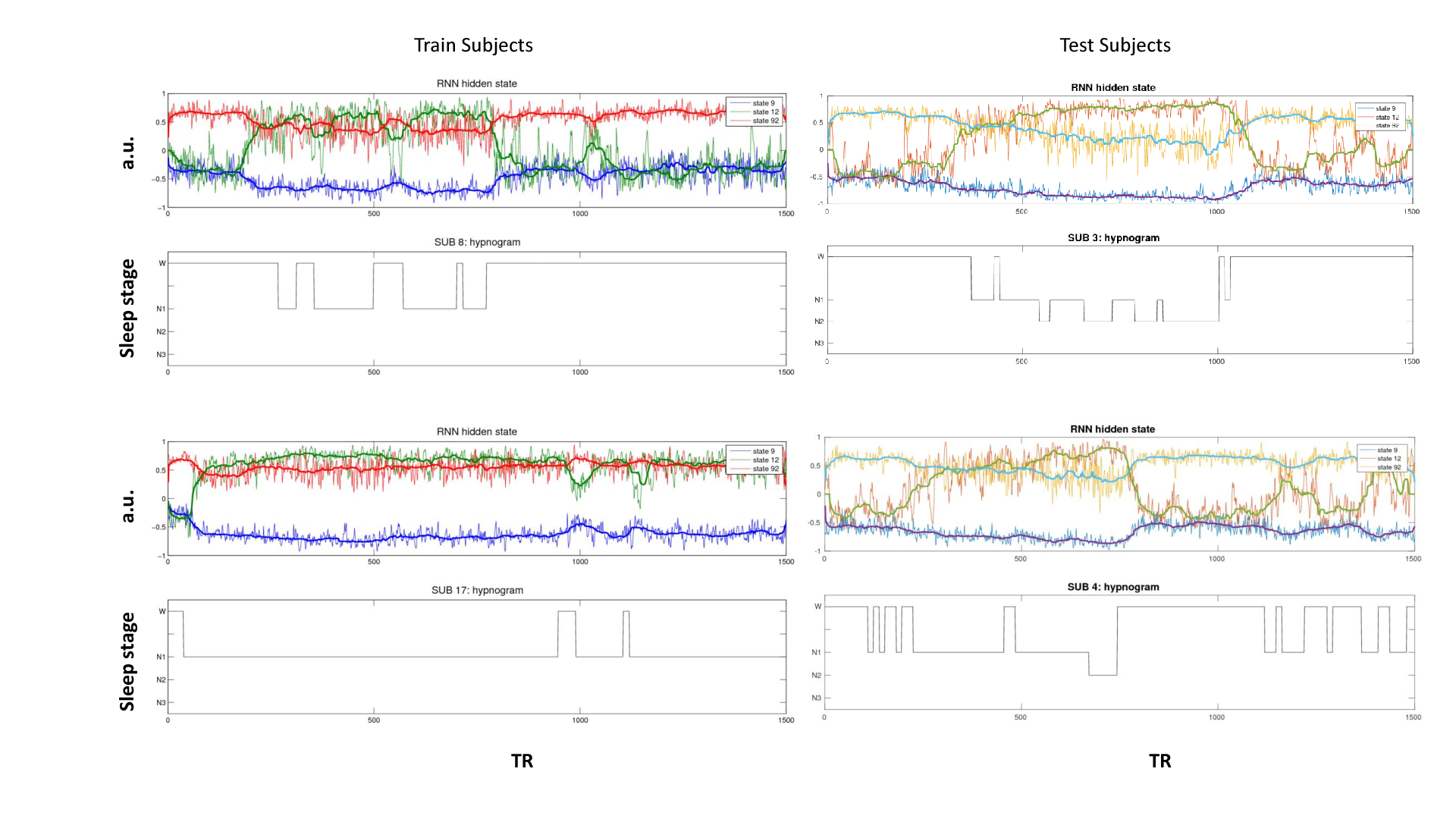}
\caption{\small
Select hidden unit time courses with the corresponding hypnogram for two subjects in the training data (left) and two test subjects (right) not used while training. Hidden states track subject neurobiological state (Wakeful (W) or N1, N2, and N3 sleep stages) just using fMRI activity. The bold lines are the median filtered activation time courses.
}
\label{fig:hypnogram}
\end{figure}

\begin{figure}[t]
\centering
\includegraphics[scale=0.9]{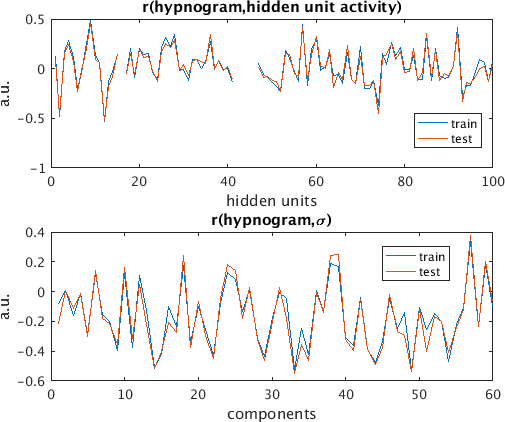}
\caption{\small
Correlation values of subject hypnogram to learned RNN model hidden units (top) and to component scale factors $\vsigma$ (bottom) for train (blue) and test (red) subjects. The same units and scale factors track subject neurobiological state as derived from EEG for both train and test cases. Note some hidden unit activations are flat and so the correlation value is empty. 
}
\label{fig:corrs}
\end{figure}

The activity of several hidden recurrent units of trained model was predictive of wakefulness across all subjects (see Figure~\ref{fig:hypnogram} for an example subject). 
The RNN hidden unit activity (bound between -1 and 1) stays at the extremes during awake state exhibiting higher standard deviation and the activity tends towards zero with lower standard deviation as the subject transitions from wakefulness to sleep. 
One-way ANOVA on the absolute mean and standard deviation of hidden unit activity by hypnogram state shows significant group differences in mean ($p \leq 10^{-29}$) and standard deviation ($p \leq 10^{-14}$). 
Subsequent posthoc t-tests reveal significant reductions in both from wakefulness and light sleep N1 state to deeper sleep stages N2 and N3 states, and also between N2 and N3 states (means:[$0.6642$ $0.6554$ $0.4558$ $0.2033$], and standard deviations: [$0.1868$ $0.1997$ $0.1567$ $0.0579$]; all these p-values $ \leq 10^{-5}$ after correcting for multiple comparisons). 
In addition, the scaling factor tended to correlate well with changes of state, as measures by correlation with a smoothed derivative of the hypnogram. 
Figure~\ref{fig:corrs} shows the correlation coefficients between RNN hidden units to subject hypnogram state, component scale factors, $\vsigma$ to subject hypnogram vector. Several hidden states show consistent correlation to hypnograms, indicating the RNN is encoding subject sleep state. 
Similarly some component scale factors also encode sleep states. Surprisingly, however, the source time courses, $\vs$, and the means, $\vmu$, did not. Finally, some component scale factors correlate somewhat consistently with changes in state across subjects. This indicates that the model is encoding changes of state in terms of uncertainty.

\section{Discussion and conclusion}
\subsection{Summary}
In this work, we demonstrate how recurrent neural networks can be used to separate conditionally independent sources analogous to independent component analysis but with the benefits of modeling temporal dynamics through recurrent parameters. 
Results show that this approach is effective for modeling both task-related and resting-state functional magnetic imaging (fMRI) data.
Using this approach, we are able to separate similar components to ICA, but having the additional benefit of directly analyzing temporal dynamics through the recurrent parameters.

Notably, in addition to finding similar maps and task-relatedness as with ICA, we are able to derive directed temporal connectivity which is task-related and group-differential, and these are derived directly from the parameters of the RNN.
In addition, for resting state data, we found that some hidden unit activity corresponded very well with wake/sleep states and that the uncertainty factor was consistent with changes of state, both of which were learned in a completely unsupervised way.

\subsection{Related work}
Our method introduces deep and nonlinear computations in time to maximum likelihood estimation independent component analysis (MLE ICA) without sacrificing the simplicity of linear relationships between source and observation.
MLE ICA has an equivalent learning objective to infomax ICA, widely used in fMRI studies, in which the sources are drawn from a factorized logistic distribution~\citep{hyvarinen2004independent}.
While the model learns a linear transformation between data and sources through the unmixing matrix, the source dynamics are encoded by a deep nonlinear transformation with recurrent structure, as represented by an RNN.
Alternative nonlinear parameterizations of the ICA transformation exist that use deep neural networks have been shown to work with fMRI data \citep{castro2016deep}.
Such approaches allow for deep and nonlinear static spatial maps and are compatible with our learning objective.
Temporal ICA as used in group ICA \citep{calhoun2009review}, like spatial ICA, does capture some temporal dynamics, but only as summaries through a one- to two-stage PCA preprocessing step.
These temporal summaries are captured and can be analyzed, however they are not learned as part of an end-to-end learning objective.
Overall, the strengths of RNN-ICA compared to these methods are the dynamics are directly learned as model parameters, which allows for richer and higher-order temporal analyses, as we showed in the previous section.

Recurrent neural networks do not typically incorporate latent variables, as this requires expensive inference.
Versions that incorporate stochastic latent variables exist, are trainable via variational methods, and working approaches for sequential data exist \citep{chung2015recurrent}.
However, these require complex inference which introduces variance into learning that may make training with fMRI data challenging.
Our method instead incorporates concepts from noiseless ICA, which reduces inference to the inverse of a generative transformation.
The consequence is that the temporal analyses are relatively simple, relying on only the tractable computation of the Jacobian of component conditional densities given the activations.

\subsection{Future work}
The RNN-ICA model provides a unique mode of analysis previously unavailable to fMRI research.
Results are encouraging, in that we were able to find both task-related and group-differentiating directed connectivity, however the broader potential of this approach is unexplored.
It is our belief that this method will expand neuroscience research that involves temporal data, leading to new and significant conclusions.

Finally, the uncertainty factor in our resting state experiments may indicate a novel application for imaging data through RNN-ICA, that is change-of-state detection.
The model we employed was simple, as was not intended to take advantage of this.
It is quite possible that further modifications could produce a model that reliably predicts change-of-state in fMRI and EEG data.

\section{Acknowledgements}
This work was supported in part by National Institutes of Health grants 2R01EB005846, P20GM103472, and \\R01EB020407 and National Science Foundation grant \#1539067

\small
\bibliography{main}

\begin{thebibliography}{}

\bibitem[Allen et~al., 2012]{allen2012b}
Allen, E.~A., Erhardt, E.~B., Wei, Y., Eichele, T., and Calhoun, V.~D. (2012).
\newblock Capturing inter-subject variability with group independent component
  analysis of {fMRI} data: a simulation study.
\newblock {\em Neuroimage}, 59(4):4141--4159.

\bibitem[Bahdanau et~al., 2014]{bahdanau2014neural}
Bahdanau, D., Cho, K., and Bengio, Y. (2014).
\newblock Neural machine translation by jointly learning to align and
  translate.
\newblock {\em arXiv preprint arXiv:1409.0473}.

\bibitem[Bell and Sejnowski, 1995]{bell1995}
Bell, A.~J. and Sejnowski, T.~J. (1995).
\newblock An information-maximization approach to blind separation and blind
  deconvolution.
\newblock {\em Neural Computation}, 7(6):1129--1159.

\bibitem[Biswal et~al., 1995]{biswal1995}
Biswal, B., Zerrin~Yetkin, F., Haughton, V.~M., and Hyde, J.~S. (1995).
\newblock Functional connectivity in the motor cortex of resting human brain
  using echo-planar {MRI}.
\newblock {\em Magnetic Resonance in Medicine}, 34(4):537--541.

\bibitem[Blondel et~al., 2008]{blondel2008fast}
Blondel, V.~D., Guillaume, J.-L., Lambiotte, R., and Lefebvre, E. (2008).
\newblock Fast unfolding of communities in large networks.
\newblock {\em Journal of statistical mechanics: theory and experiment},
  2008(10):P10008.

\bibitem[Calhoun et~al., 2001a]{calhoun2001spatial}
Calhoun, V., Adali, T., Pearlson, G., and Pekar, J. (2001a).
\newblock Spatial and temporal independent component analysis of functional mri
  data containing a pair of task-related waveforms.
\newblock {\em Human brain mapping}, 13(1):43--53.

\bibitem[Calhoun and Adali, 2012]{calhoun2012multisubject}
Calhoun, V.~D. and Adali, T. (2012).
\newblock Multisubject independent component analysis of fmri: a decade of
  intrinsic networks, default mode, and neurodiagnostic discovery.
\newblock {\em IEEE reviews in biomedical engineering}, 5:60--73.

\bibitem[Calhoun et~al., 2001b]{vince2001}
Calhoun, V.~D., Adali, T., Pearlson, G.~D., and Pekar, J.~J. (2001b).
\newblock A method for making group inferences from functional {MRI} data using
  independent component analysis.
\newblock {\em Human Brain Mapping}, 14.

\bibitem[Calhoun et~al., 2008]{vince2008}
Calhoun, V.~D., Kiehl, K.~A., and Pearlson, G.~D. (2008).
\newblock Modulation of temporally coherent brain networks estimated using
  {ICA} at rest and during cognitive tasks.
\newblock {\em Human Brain Mapping}, 29(7):828--838.

\bibitem[Calhoun et~al., 2009]{calhoun2009review}
Calhoun, V.~D., Liu, J., and Adal{\i}, T. (2009).
\newblock A review of group ica for fmri data and ica for joint inference of
  imaging, genetic, and erp data.
\newblock {\em Neuroimage}, 45(1):S163--S172.

\bibitem[Castro et~al., 2016]{castro2016deep}
Castro, E., Hjelm, R.~D., Plis, S., Dihn, L., Turner, J., and Calhoun, V.
  (2016).
\newblock Deep independence network analysis of structural brain imaging:
  Application to schizophrenia.
\newblock {\em IEEE}.

\bibitem[Cho et~al., 2014]{cho2014learning}
Cho, K., Van~Merri{\"e}nboer, B., Gulcehre, C., Bahdanau, D., Bougares, F.,
  Schwenk, H., and Bengio, Y. (2014).
\newblock Learning phrase representations using rnn encoder-decoder for
  statistical machine translation.
\newblock {\em arXiv preprint arXiv:1406.1078}.

\bibitem[Chung et~al., 2015]{chung2015recurrent}
Chung, J., Kastner, K., Dinh, L., Goel, K., Courville, A.~C., and Bengio, Y.
  (2015).
\newblock A recurrent latent variable model for sequential data.
\newblock In {\em Advances in neural information processing systems}, pages
  2962--2970.

\bibitem[Cox, 1996]{cox1996afni}
Cox, R.~W. (1996).
\newblock Afni: software for analysis and visualization of functional magnetic
  resonance neuroimages.
\newblock {\em Computers and Biomedical research}, 29(3):162--173.

\bibitem[Damaraju et~al., 2014]{damaraju2014dynamic}
Damaraju, E., Allen, E.~A., Belger, A., Ford, J., McEwen, S., Mathalon, D.,
  Mueller, B., Pearlson, G., Potkin, S., Preda, A., et~al. (2014).
\newblock Dynamic functional connectivity analysis reveals transient states of
  dysconnectivity in schizophrenia.
\newblock {\em NeuroImage: Clinical}, 5:298--308.

\bibitem[Damoiseaux et~al., 2006]{damoiseaux2006}
Damoiseaux, J., Rombouts, S., Barkhof, F., Scheltens, P., Stam, C., Smith,
  S.~M., and Beckmann, C. (2006).
\newblock Consistent resting-state networks across healthy subjects.
\newblock {\em Proceedings of the National Academy of Sciences},
  103(37):13848--13853.

\bibitem[Doucet et~al., 2001]{doucet2001introduction}
Doucet, A., De~Freitas, N., and Gordon, N. (2001).
\newblock An introduction to sequential monte carlo methods.
\newblock In {\em Sequential Monte Carlo methods in practice}, pages 3--14.
  Springer.

\bibitem[Erhardt et~al., 2012]{erhardt2012simtb}
Erhardt, E.~B., Allen, E.~A., Wei, Y., Eichele, T., and Calhoun, V.~D. (2012).
\newblock Simtb, a simulation toolbox for fmri data under a model of
  spatiotemporal separability.
\newblock {\em Neuroimage}, 59(4):4160--4167.

\bibitem[Fu et~al., 2014]{fu2014blind}
Fu, G.-S., Phlypo, R., Anderson, M., Li, X.-L., et~al. (2014).
\newblock Blind source separation by entropy rate minimization.
\newblock {\em IEEE Transactions on Signal Processing}, 62(16):4245--4255.

\bibitem[Graves, 2013]{graves2013generating}
Graves, A. (2013).
\newblock Generating sequences with recurrent neural networks.
\newblock {\em arXiv preprint arXiv:1308.0850}.

\bibitem[G{\"u}{\c{c}}l{\"u} and van Gerven, 2017]{gucclu2017modeling}
G{\"u}{\c{c}}l{\"u}, U. and van Gerven, M.~A. (2017).
\newblock Modeling the dynamics of human brain activity with recurrent neural
  networks.
\newblock {\em Frontiers in computational neuroscience}, 11.

\bibitem[Hinton, 2012]{Hinton-Coursera2012}
Hinton, G. (2012).
\newblock Neural networks for machine learning.
\newblock Coursera, video lectures.

\bibitem[Hjelm et~al., 2014]{hjelm2014restricted}
Hjelm, R.~D., Calhoun, V.~D., Salakhutdinov, R., Allen, E.~A., Adali, T., and
  Plis, S.~M. (2014).
\newblock Restricted boltzmann machines for neuroimaging: an application in
  identifying intrinsic networks.
\newblock {\em NeuroImage}, 96:245--260.

\bibitem[Hochreiter and Schmidhuber, 1997]{hochreiter1997long}
Hochreiter, S. and Schmidhuber, J. (1997).
\newblock Long short-term memory.
\newblock {\em Neural computation}, 9(8):1735--1780.

\bibitem[Hyv{\"a}rinen et~al., 2004]{hyvarinen2004independent}
Hyv{\"a}rinen, A., Karhunen, J., and Oja, E. (2004).
\newblock {\em Independent component analysis}, volume~46.
\newblock John Wiley \& Sons.

\bibitem[Hyv{\"a}rinen and Oja, 2000]{hyvarinen2000independent}
Hyv{\"a}rinen, A. and Oja, E. (2000).
\newblock Independent component analysis: algorithms and applications.
\newblock {\em Neural networks}, 13(4):411--430.

\bibitem[Jafri et~al., 2008]{jafri2008method}
Jafri, M.~J., Pearlson, G.~D., Stevens, M., and Calhoun, V.~D. (2008).
\newblock A method for functional network connectivity among spatially
  independent resting-state components in schizophrenia.
\newblock {\em Neuroimage}, 39(4):1666--1681.

\bibitem[Kim et~al., 2008]{kim2008hybrid}
Kim, D., Burge, J., Lane, T., Pearlson, G.~D., Kiehl, K.~A., and Calhoun, V.~D.
  (2008).
\newblock Hybrid ica--bayesian network approach reveals distinct effective
  connectivity differences in schizophrenia.
\newblock {\em Neuroimage}, 42(4):1560--1568.

\bibitem[Lehmann et~al., 2017]{LEHMANN2017635}
Lehmann, B., White, S., Henson, R., Cam-CAN, and Geerligs, L. (2017).
\newblock Assessing dynamic functional connectivity in heterogeneous samples.
\newblock {\em NeuroImage}, 157:635 -- 647.

\bibitem[Papademetris et~al., 2006]{papademetris2006bioimage}
Papademetris, X., Jackowski, M.~P., Rajeevan, N., DiStasio, M., Okuda, H.,
  Constable, R.~T., and Staib, L.~H. (2006).
\newblock Bioimage suite: An integrated medical image analysis suite: An
  update.
\newblock {\em The insight journal}, 2006:209.

\bibitem[Plis et~al., 2013]{plis2013deep}
Plis, S.~M., Hjelm, R.~D., Salakhutdinov, R., and Calhoun, V.~D. (2013).
\newblock Deep learning for neuroimaging: a validation study.
\newblock {\em Human Brain Mapping}.

\bibitem[Shen et~al., 2013]{shen2013groupwise}
Shen, X., Tokoglu, F., Papademetris, X., and Constable, R.~T. (2013).
\newblock Groupwise whole-brain parcellation from resting-state fmri data for
  network node identification.
\newblock {\em Neuroimage}, 82:403--415.

\bibitem[Smith et~al., 2009]{smith2009}
Smith, S.~M., Fox, P.~T., Miller, K.~L., Glahn, D.~C., Fox, P.~M., Mackay,
  C.~E., Filippini, N., Watkins, K.~E., Toro, R., Laird, A.~R., et~al. (2009).
\newblock Correspondence of the brain's functional architecture during
  activation and rest.
\newblock {\em Proceedings of the National Academy of Sciences},
  106(31):13040--13045.

\bibitem[Sutskever et~al., 2014]{sutskever2014sequence}
Sutskever, I., Vinyals, O., and Le, Q.~V. (2014).
\newblock Sequence to sequence learning with neural networks.
\newblock In {\em Advances in neural information processing systems}, pages
  3104--3112.

\bibitem[Swanson et~al., 2011]{swanson2010}
Swanson, N., Eichele, T., Pearlson, G., Kiehl, K., Yu, Q., and Calhoun, V.~D.
  (2011).
\newblock Lateral differences in the default mode network in healthy controls
  and patients with schizophrenia.
\newblock {\em Human Brain Mapping}, 32(4):654--664.

\bibitem[Tagliazucchi et~al., 2012]{TAGLIAZUCCHI201263}
Tagliazucchi, E., von Wegner, F., Morzelewski, A., Borisov, S., Jahnke, K., and
  Laufs, H. (2012).
\newblock Automatic sleep staging using fmri functional connectivity data.
\newblock {\em NeuroImage}, 63(1):63 -- 72.

\bibitem[Van Den~Oord et~al., 2016]{van2016wavenet}
Van Den~Oord, A., Dieleman, S., Zen, H., Simonyan, K., Vinyals, O., Graves, A.,
  Kalchbrenner, N., Senior, A., and Kavukcuoglu, K. (2016).
\newblock Wavenet: A generative model for raw audio.
\newblock {\em arXiv preprint arXiv:1609.03499}.

\bibitem[Vergara et~al., 2017]{vergara2017effect}
Vergara, V.~M., Mayer, A.~R., Damaraju, E., and Calhoun, V.~D. (2017).
\newblock The effect of preprocessing in dynamic functional network
  connectivity used to classify mild traumatic brain injury.
\newblock {\em Brain and behavior}, 7(10).

\bibitem[Zuo et~al., 2010]{zuo2010}
Zuo, X.-N., Kelly, C., Adelstein, J.~S., Klein, D.~F., Castellanos, F.~X., and
  Milham, M.~P. (2010).
\newblock Reliable intrinsic connectivity networks: test--retest evaluation
  using {ICA} and dual regression approach.
\newblock {\em Neuroimage}, 49(3):2163--2177.

\end{thebibliography}
\bibliographystyle{apalike}

\end{document}